\documentclass[sigconf]{acmart}

\usepackage{todonotes}

\renewcommand\footnotetextcopyrightpermission[1]{} 
\acmYear{2025}
\acmISBN{978-1-4503-XXXX-X}
\acmDOI{10.1145/XXXXXXX.XXXXXXX}

\acmConference[Preprint]{Under Review}{2025}{}

\usepackage{tcolorbox}
\definecolor{silver}{RGB}{230,230,230}

\usepackage{tcolorbox}
\tcbset{
  colback=gray!3, 
  colframe=gray!60, 
  fonttitle=\bfseries\small,
  coltitle=black,
  boxsep=1mm,
  arc=1.5mm,
  top=1mm,
  bottom=1mm,
  left=1mm,
  right=1mm,
  width=0.5\textwidth,
  sharp corners=southwest 
}

\author{Kasu Sai Kartheek Reddy\textsuperscript{\textdagger}}
\affiliation{
  \city{IIIT Dharwad}
  \country{India}
}
\email{saikartheekreddykasu@gmail.com}

\author{Shankar Biradar}
\affiliation{
  \city{MIT Manipal}
  \country{India}
}
\email{shankar.biradar@manipal.edu}

\author{Sunil Saumya}
\affiliation{
  \city{IIIT Dharwad}
  \country{India}
}
\email{sunil.saumya@iiitdwd.ac.in}

\thanks{\textsuperscript{\textdagger}Corresponding author}

\begin{document}

\title{Deceptive Humor: A Synthetic Multilingual Benchmark Dataset for Bridging Fabricated Claims with Humorous Content}



\begin{abstract}
In the evolving landscape of online discourse, misinformation increasingly adopts humorous tones to evade detection and gain traction. This work introduces Deceptive Humor as a novel research direction, emphasizing how false narratives, when coated in humor, can become more difficult to detect and more likely to spread. To support research in this space, we present the Deceptive Humor Dataset (DHD) a collection of humor-infused comments derived from fabricated claims using the ChatGPT-4o model. Each entry is labeled with a Satire Level (from 1 for subtle satire to 3 for overt satire) and categorized into five humor types: Dark Humor, Irony, Social Commentary, Wordplay, and Absurdity. The dataset spans English, Telugu, Hindi, Kannada, Tamil, and their code-mixed forms, making it a valuable resource for multilingual analysis. DHD offers a structured foundation for understanding how humor can serve as a vehicle for the propagation of misinformation, subtly enhancing its reach and impact. Strong baselines are established to encourage further research and model development in this emerging area.

\end{abstract}






\maketitle
\vspace{-15pt}
\textcolor{red}{\textbf{Caution:} This paper includes LLM-generated data on fabricated humor that may unintentionally offend the readers.}
\vspace{-5pt}
\section{Introduction}
\begin{figure}[!ht] 
    \centering
    \includegraphics[width=0.40\textwidth, height=0.18\textheight]{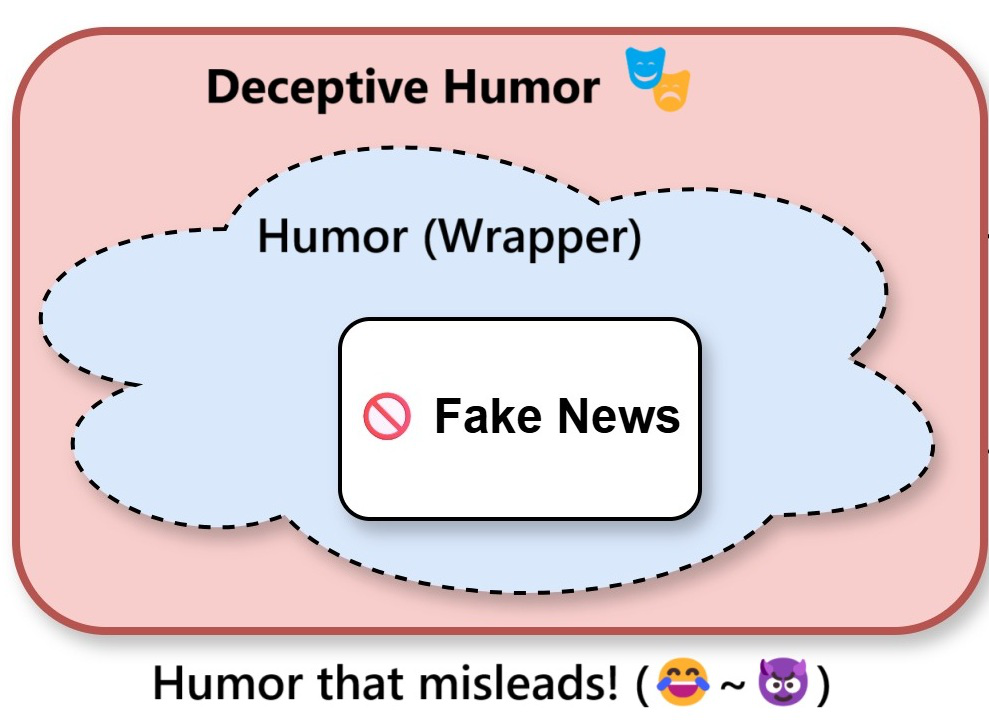} 
    \caption{Unmasking Deceptive Humor: Fake claims are embedded in humor, making them more engaging and harder to detect. The figure shows how humor serves as a wrapper.}
    \label{fig:Deceptive Humor}
\end{figure}

In today’s online world, deceptive humor is emerging as a powerful and complex form of communication. At first glance, these humorous comments seem harmless and entertaining, often making people laugh or smile. However, beneath the surface, they carry hidden falsehoods and misinformation. Because humor lowers our guard, repeated exposure to such content can subtly influence the subconscious mind, leading individuals to accept these misleading ideas without realizing it. This makes deceptive humor a double-edged sword: while it entertains, it also becomes a dangerous tool that spreads false narratives under the cover of comedy. Unlike traditional humor, which aims simply to entertain, deceptive humor deliberately masks fabricated news in playful tones, making misinformation harder to detect and easier to spread, as shown in \textcolor{blue}{\autoref{fig:Deceptive Humor}}. Understanding this unique blend of humor and deception is critical, as it reveals how misinformation can silently propagate through social platforms without raising immediate suspicion or resistance.

Consider the fabricated claim: “\textbf{Ch*na is spreading COVID as a bioweapon.}” On the surface, humorous comments related to this claim may seem lighthearted or simply playful, not drawing direct attention to the false narrative itself. However, when these comments are examined collectively and in depth, the underlying misinformation linking them becomes evident. This highlights how deceptive humor can subtly embed and reinforce fabricated claims without explicitly stating them. \textcolor{blue}{\autoref{traditional vs deceptive}} presents examples contrasting traditional humorous comments and deceptive humor comments that covertly propagate the fabricated narrative about Ch*na. This comparison demonstrates the nuanced way in which deceptive humor operates, disguising falsehoods within humor, making misinformation more difficult to detect and more likely to spread\footnote{
\href{https://sai-kartheek-reddy.github.io/Deceptive-Humor-Web/}{\textcolor{blue}{Project Website}}
}.
\vspace{-5pt}

\begin{table}[!ht]
\resizebox{0.48\textwidth}{!}{
\begin{tabular}{l}
\hline
\textbf{Traditional Humorous Comments} \\
\hline
\textit{Ch*na products usually don’t last long.} \\
\textit{My phone broke in a week, must be made in Ch*na.} \\
\hline
\textbf{Deceptive Humor Comments} \\
\hline
\textit{Ch*na products usually don’t last long, except this COVID thing.} \\
\textit{Guess Ch*na finally made something that went global and stayed.} \\
\hline
\end{tabular}
}
\caption{Understanding the difference between traditional humor and Deceptive humor.}
\label{traditional vs deceptive}
\end{table}

\vspace{-22pt}

While the prior research has explored related areas, such as Faux Hate\cite{biradar2024faux}, deceptive humor presents a distinct challenge. Unlike explicit hate speech, which users may hesitate to share, deceptive humor often appears harmless and easily propagates, making it a more insidious carrier of misinformation. Recognizing this gap, we introduce DHD, a structured resource to facilitate the systematic study of deceptive humor and its role in misinformation propagation.

\textbf{Key Contributions:}
\begin{itemize}
    \item \textbf{Dataset Contribution:} We introduce the Deceptive Humor Dataset (DHD), a novel multilingual resource for studying how humor veils misinformation, supported by detailed satire-level and humor-type labels.
    \item \textbf{Technical Contribution:} We establish strong baselines by evaluating the dataset with diverse pre-trained language models (PLMs), providing valuable benchmarks to guide and facilitate future research in fact-aware humor detection.
\end{itemize}
The remainder of the article is organized as follows: \textcolor{blue}{\autoref{lit}} reviews existing work on humor and misinformation. \textcolor{blue}{\autoref{method}} details the dataset construction process and baseline methods for deceptive humor detection. \textcolor{blue}{\autoref{results}} presents and analyzes the experimental findings and \textcolor{blue}{\autoref{humaneval}} present the alignment between human and machine labeling. Finally, \textcolor{blue}{\autoref{conclusion}} summarizes key takeaways and outlines future research directions.


\section{Literature Review}\label{lit}

Existing research treats humor and misinformation as separate domains, yet their intersection remains largely unexplored. From a psychological and social standpoint, the Interpersonal Humor Deception Model (IHDM) suggests that humor can either reduce self-centered deception and build trust or, if poorly executed, raise suspicions and undermine credibility\citep{gaspar2023laughter}. Similarly, humor is widely used in advertising to mask deceptive practices, with 73.5\% of humorous advertisements containing misleading elements that obscure unethical messaging\citep{shabbir2007use}. These findings highlight humor’s dual nature, it can both expose and conceal deception, making it an effective yet ethically complex tool.

From a computational perspective, humor detection in NLP has largely focused on sarcasm \cite{joshi2015harnessing}, irony\cite{van2018semeval}, and satire\cite{rubin2016fake}, but these studies do not address humor generated from fabricated claims. While humor’s role in misinformation detection has been acknowledged\cite{zhou2020survey}, prior work treats humor and misinformation as distinct problems, lacking a framework to analyze how humor itself can be a carrier of deceptive content. Additionally, research on fact-checking and misinformation detection\cite{thorne2018fever,bhardwaj2020hostility} has primarily focused on textual veracity, without considering how humor can distort factual claims, making detection even more complex. Current humor datasets\cite{hossain2019president} focus on linguistic features rather than fact-aware humor, limiting their applicability to deceptive humor detection.

\section{Dataset Development}\label{method}
In this section, we describe the process of selecting fabricated claims and generating the Deceptive Humor Dataset (DHD). Using ChatGPT-4o, we create humor-infused comments across multiple languages, ensuring diversity in satire and linguistic variations. We also highlight the role of synthetic data in advancing AI, emphasizing its importance in training robust models and addressing data scarcity in multilingual settings.
\subsection{Selection of Fake Claims}
The first step in data acquisition involves identifying various topics for data collection. To ensure diversity in the collected data, the authors have selected a range of topics, including entertainment, politics, finance, sports, religion, and health. Following the selection of topics, the next step is to identify fake narratives associated with these topics. These fake narratives are systematically scraped from well-known fact-checking websites such as AltNews\footnote{https://www.altnews.in/}, Boom FactCheck\footnote{https://www.boomlive.in/fact-check}, FactChecker\footnote{https://www.factchecker.in/fact-check}, and FACTLY\footnote{https://factly.in/}. This approach ensures the reliability and relevance of the data collected for analysis.

\subsection{Generation of Deceptive Humor corpus}
In this section, we describe the process of generating the DHD. The dataset is constructed using the ChatGPT-4o \cite{hurst2024gpt} model. We adopt synthetic data generation due to the inherently complex nature of deceptive humor. While deceptive comments do occur frequently in real-world settings, they are often difficult to detect and collect reliably at scale because they tend to blend subtly into regular discourse and require contextual or factual background to identify. To address these challenges, we leverage the controllability and scalability of LLMs to generate high-quality, humor-infused comments grounded in fabricated claims. This approach allows us to ensure consistency across multiple languages and maintain diversity in humor styles and linguistic variations.

Generating humorous content that incorporates deception is particularly challenging, as it requires balancing satire with subtle misinformation. To explore the most effective tools for this task, we evaluated several state-of-the-art generative models, including Gemini \cite{team2023gemini}, LLaMA \cite{touvron2023llama}, Claude, and ChatGPT. While Gemini and LLaMA perform reasonably well for English, they often produce ungrammatical or incoherent results in Indic languages. Claude, meanwhile, frequently declines requests involving humor and deception due to its usage policies. After a thorough evaluation, we find that ChatGPT-4o consistently generates coherent, contextually appropriate, and humorous comments across languages, making it the most suitable choice for constructing the Deceptive Humor Dataset.




The DHD corpus is generated using ChatGPT-4o with carefully structured prompts aimed at producing natural and human-like humor. To ensure the content remained appropriate and high-quality, language experts aged 18 or above reviewed and filtered the generated outputs. This human-in-the-loop process ensured the humor was engaging while avoiding content that could negatively impact younger or sensitive audiences. Details of the structured prompting approach are provided in the \href{https://github.com/Sai-Kartheek-Reddy/Deceptive-Humor-Web/blob/main/Metainformation%20and%20Configurations/Prompt.txt}{\textcolor{blue}{prompt design}}

\begin{figure}[!ht] 
    \centering
    \includegraphics[width=0.48\textwidth, height=0.2\textheight]{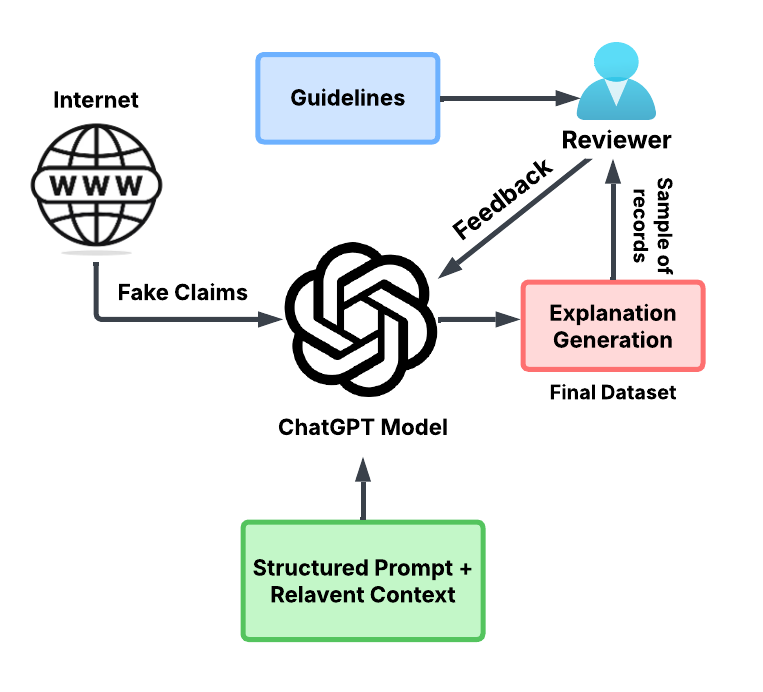} 
    \caption{Flowchart for the DHD Data Generation}
    \label{fig:data_flow}
\end{figure}

We enhance data complexity by combining multiple claims and tasking the model with generating comments, resulting in more nuanced samples. If the quality is lacking, we refine the output using feedback through instructions or examples, and by providing relevant context to make the model aware of the background of the claim, thereby improving the human-likeness of the text, as shown in \textcolor{blue}{\autoref{fig:data_flow}}.

\noindent \textbf{Role of Synthetic Data in Advancing the AI systems:} \\
The proposed DHD is synthetically generated using the ChatGPT-4o model. A common critique of synthetic data is that PLMs struggle to capture patterns representative of human-generated text. While this concern has some merit, it is important to recognize that human annotations themselves are influenced by inherent biases shaped by individual mental models \cite{gautam2024blind}. The role of synthetic data in AI research has grown substantially, with top institutions like Hugging Face and various companies actively developing synthetic data generators\footnote{https://huggingface.co/blog/synthetic-data-generator} to support this effort. Notably, the Phi-4 model, a SOTA open model, incorporates synthetic data as a core component of its training regimen, underscoring its practical value in advancing AI capabilities.

Recent work across leading AI research venues further validates synthetic data’s critical role in improving model generalization, addressing data scarcity, and mitigating annotation biases. For instance, Google DeepMind’s comprehensive study outlines best practices and challenges in synthetic data generation, highlighting its potential to enhance model robustness and fairness \cite{liu2024best}. In multimodal learning, synthetic data has been demonstrated to boost unsupervised visual representation learning by generating effective training samples and improving data efficiency \cite{wu2023synthetic}. Additionally, synthetic data-driven self-training methods have shown promise in low-resource natural language processing tasks such as relation extraction, effectively overcoming domain adaptation challenges \cite{xu2023s2ynre}. These advancements position synthetic data not merely as a workaround for limited human annotations but as a transformative tool that drives innovation and broader applicability in AI. In this light, synthetic data provides an essential foundation for our work on Deceptive Humor detection and enables future research progress in this emerging domain.
\subsection{Dataset Description}
The proposed DHD consists of 9,000 synthetically generated humorous comments, carefully curated to ensure linguistic diversity and humor variation. The dataset is split into 7,200 comments for training, 900 for validation, and 900 for testing as shown in \textcolor{blue}{\autoref{tab:tab1}}. Each comment is labelled with a satire level ranging from 1 to 3, where 1 represents subtle satire and 3 denotes highly exaggerated satire. Additionally, every comment is assigned a humor attribute from one of the five predefined categories: Irony, Absurdity, Social Commentary, Dark Humor, and Wordplay.
\begin{table}[!ht]
\resizebox{0.48\textwidth}{!}{
\begin{tabular}{lccc}
\hline
\textbf{Statistic} & \textbf{Train} & \textbf{Validation} & \textbf{Test} \\
\hline
\textbf{Total Samples} & 7,200 & 900 & 900 \\
\hline
\multicolumn{4}{l}{\textbf{Satire Level Distribution}} \\
\hline
Low Satire & 2,083 & 276 & 329 \\
Moderate Satire & 3,138 & 382 & 271 \\
High Satire & 1,990 & 242 & 300 \\
\hline
\multicolumn{4}{l}{\textbf{Humor Attribute Distribution}} \\
\hline
Irony & 2,200 & 282 & 245 \\
Absurdity & 1,661 & 180 & 258 \\
Social Commentary & 1,215 & 155 & 102 \\
Dark Humor & 1,196 & 136 & 156 \\
Wordplay & 1,039 & 147 & 139 \\
\hline
\end{tabular}
}
\caption{DHD Distribution}
\label{tab:tab1}
\end{table}

A key aspect of the DHD is its linguistic diversity. Along with English, it includes comments in four major Indic languages: Telugu, Hindi, Kannada, and Tamil, along with their code-mixed versions. This ensures a rich and varied dataset that captures the nuances of humor across multiple languages and cultural contexts. The structured labeling enables a comprehensive analysis of humor in NLP systems, fostering advancements in computational humor understanding, particularly in multilingual and code-mixed settings. A detailed description of the dataset is presented in \textcolor{blue}{\autoref{tab:tab3}}\\

\noindent\textbf{Description of the Labels:}

\noindent
\noindent
\textbf{Satire Level:} This label quantifies the degree of satire in the generated comment.

\begin{itemize}
    \item \textbf{Low Satire:} The humor is subtle and lightly satirical, often resembling real-world statements with a mild twist.
    \item \textbf{Moderate Satire:} The humor is more evident, incorporating exaggeration and sarcasm while maintaining a balance between reality and absurdity.
    \item \textbf{High Satire:} The humor is strongly exaggerated and overtly satirical, often making use of extreme irony or absurd distortions of reality.
\end{itemize}

\vspace{1em}

\noindent
\textbf{Humor Attribute:} This label categorizes the type of humor used in the comment.

\begin{itemize}
    \item \textbf{Irony}\footnote{\href{https://en.wikipedia.org/wiki/Irony}{Irony Wikipedia}}: A form of humor where the intended meaning contrasts sharply with the literal meaning, often exposing contradictions or unexpected outcomes.
    
    \item \textbf{Absurdity}\footnote{\href{https://en.wikipedia.org/wiki/Surreal_humour}{Absurdity Wikipedia}}: Humor that thrives on exaggeration, illogical scenarios, or unrealistic premises to create an amusing effect.
    
    \item \textbf{Social Commentary}\footnote{\href{https://en.wikipedia.org/wiki/Social_commentary}{Social Commentary Wikipedia}}: Humor that critiques, mocks, or highlights societal or cultural issues, often with a satirical or thought-provoking angle.
    
    \item \textbf{Dark Humor}\footnote{\href{https://en.wikipedia.org/wiki/Black_comedy}{Dark Humor Wikipedia}}: Humor that deals with morbid, taboo, or controversial topics in a way that might be unsettling but still amusing.
    
    \item \textbf{Wordplay}\footnote{\href{https://en.wikipedia.org/wiki/Word_play}{Wordplay Wikipedia}}: Humor that relies on clever linguistic constructs, including puns, double meanings, and phonetic playfulness.
\end{itemize}

\begin{table*}[!ht]
\centering
\resizebox{\textwidth}{!}{
\begin{tabular}{l cccccccccc}
\hline
\textbf{Language} & \textbf{Total Records} & \multicolumn{3}{c|}{\textbf{Satire Level}} & \multicolumn{5}{c}{\textbf{Humor Attribute}} \\
\cline{3-5} \cline{6-10}
 & & \textbf{1} & \textbf{2} & \textbf{3} & \textbf{Absurdity} & \textbf{Dark Humor} & \textbf{Irony} & \textbf{Social Commentary} & \textbf{Wordplay} \\
\hline
\multicolumn{10}{l}{\textbf{Train Data}} \\
\hline
English & 881 & 188 & 322 & 371 & 259 & 149 & 292 & 101 & 80 \\
Telugu* & 1665 & 495 & 698 & 472 & 372 & 276 & 552 & 274 & 191 \\
Hindi* & 1626 & 481 & 735 & 410 & 284 & 205 & 510 & 316 & 311 \\
Kannada* & 1539 & 461 & 674 & 404 & 397 & 265 & 384 & 285 & 208 \\
Tamil* & 1500 & 458 & 709 & 333 & 349 & 201 & 462 & 239 & 249 \\
\hline
\multicolumn{10}{l}{\textbf{Validation Data}} \\
\hline
English & 114 & 25 & 37 & 52 & 22 & 25 & 37 & 14 & 16 \\
Telugu* & 221 & 68 & 99 & 54 & 47 & 33 & 67 & 35 & 39 \\
Hindi* & 195 & 59 & 86 & 50 & 35 & 26 & 52 & 34 & 48 \\
Kannada* & 198 & 67 & 86 & 45 & 47 & 30 & 62 & 39 & 20 \\
Tamil* & 172 & 57 & 74 & 41 & 29 & 22 & 64 & 33 & 24 \\
\hline
\multicolumn{10}{l}{\textbf{Test Data (Human Annotated)}} \\
\hline
English & 104 & 33 & 36 & 35 & 31 & 15 & 35 & 7 & 16 \\
Telugu* & 204 & 111 & 65 & 28 & 70 & 21 & 57 & 21 & 35 \\
Hindi* & 207 & 36 & 56 & 115 & 35 & 57 & 71 & 14 & 31 \\
Kannada* & 189 & 67 & 40 & 82 & 62 & 40 & 39 & 20 & 28 \\
Tamil* & 196 & 82 & 74 & 40 & 61 & 23 & 43 & 40 & 29 \\
\hline
\end{tabular}
}
\caption{Distribution of labels Across Languages. (* indicates Indic languages along with their code-mixed variants.)}
\label{tab:tab3}
\end{table*}

\section{Experimental Setup and Results}\label{results}

We evaluated a range of model architectures—Encoder-Only, Encoder-Decoder, and LLMs \cite{brown2020language}, across Zero-Shot, Few-Shot, and QLoRA-based\cite{dettmers2024qlora} fine-tuning settings, targeting both Satire Level classification and Humor Attribute prediction. Encoder-Only models, particularly mBERT and BERT \cite{kenton2019bert}, consistently outperformed others, with mBERT excelling in Satire Level classification and BERT leading in Humor Attribute prediction. In contrast, LLMs in Zero-Shot and Few-Shot settings exhibited inconsistent behavior, often failing to predict certain labels. LLMs, even after QLoRA fine-tuning, lag behind Encoder-Only and Encoder-Decoder models on the test set, as evident in \textcolor{blue}{\autoref{tab:baseline_results}}.

\begin{table}[!ht]
\resizebox{0.48\textwidth}{!}{
\begin{tabular}{lccc|cc}
\hline
\textbf{Language} & \multicolumn{3}{c|}{\textbf{Satire Level (mBERT)}} & \multicolumn{2}{c}{\textbf{Humor Attribute (mBART)}} \\
 & \textbf{Acc} & \textbf{F1} & \textbf{Pear} & \textbf{Acc} & \textbf{F1} \\
\hline
English & \textbf{44.23} & 42.96 & 18.97 & 34.62  & 33.99  \\
Telugu* & 44.12 & \textbf{45.52} & 27.84 & 31.37 & 30.75 \\
Hindi* & 43.48 & 45.22 & \textbf{42.12} & \textbf{34.78} & \textbf{35.39} \\
Kannada* & 38.62 & 39.88 & 23.01 & 25.93 & 25.45 \\
Tamil* & 41.33 & 41.28 & 15.13 & 27.55 & 25.75 \\
\hline
\end{tabular}
}
\caption{Language-wise performance on Satire Level and Humor Attribute tasks. Metrics include Accuracy (Acc), F1 Score (F1), and Pearson Correlation (Pear)}
\label{tab:langwise_perf}
\end{table}

\vspace{-20pt}
These findings underscore the challenges LLMs face with nuanced tasks like deceptive humor classification, which requires deep contextual and cultural understanding. The subtlety and ambiguity of deceptive humor often cause misclassifications or omission of certain classes, highlighting the need for further advancements. Prior work, such as the Memotion analysis task\cite{sharma-etal-2020-semeval}, has also shown that humor detection becomes increasingly difficult when fine-grained classification is involved, especially with limited modalities. Consistent with those findings, our experiments reveal that when only the text modality is used, performance significantly drops, reinforcing the open nature of this research problem. \textcolor{blue}{\autoref{tab:langwise_perf}}, presents the best-performing model's results broken down by language, providing a clearer picture of performance variations across different linguistic contexts. A detailed error analysis and discussion of challenging samples are provided in \textcolor{blue}{\autoref{error_analysis}}, offering insights into current limitations and avenues for future research.\\


Our findings reveal a fundamental challenge: existing models, which perform well on humor detection and misinformation classification individually, struggle significantly when dealing with deceptive humor. This is because deceptive humor is not merely about understanding humor; it also requires fact verification and intent recognition, which traditional humor detection models lack. While humor classification has been explored using datasets such as SemEval-2017 Task 6 on Humor Detection \citep{potash2017semeval}, Humicroedit \citep{hossain2019president}, these datasets primarily focus on linguistic humor rather than humor derived from fabricated claims. Similarly, misinformation detection datasets such as FEVER \citep{thorne2018fever}, LIAR \citep{wang2017liar}, and Hostile Dataset\citep{bhardwaj2020hostility} focus on textual veracity but do not account for the nuances of humor distorting false narratives. Although synthetic humor datasets have been introduced, such as the Unfun dataset \citep{horvitz2024getting}, which edits humorous texts to make them non-humorous for improved humor detection, their primary focus is on humor manipulation rather than humor intertwined with deception. This highlights the increased complexity of deceptive humor, which combines linguistic ambiguity, misinformation, and humor-specific reasoning, making it a more challenging task than standard humor detection or fact-checking alone.\footnote{ColBERT Humor Detection Dataset: \url{https://huggingface.co/datasets/CreativeLang/ColBERT_Humor_Detection}}

Unlike prior work that treats humor and misinformation as separate problems, our findings highlight how humor itself can actively propagate misinformation, further complicating detection. This introduces a fundamentally new challenge that requires both linguistic and factual reasoning, where existing models struggle. Our results in \autoref{tab:baseline_results} confirm this gap even fine-tuned transformer models and LLMs fail to generalize effectively across different humor attributes and satire levels in the proposed DHD. These limitations emphasize the urgent need for novel approaches that integrate both humor understanding and misinformation detection to effectively handle deceptive humor.

\begin{table*}[!ht]
\centering
\resizebox{\textwidth}{!}{
\begin{tabular}{l ccccccccc}
\hline
\textbf{Model} & \multicolumn{5}{c|}{\textbf{Satire Level}} & \multicolumn{4}{c
}{\textbf{Humor Attribute}} \\
\cline{2-10}
& \textbf{Accuracy} & \textbf{F1-Score} & \textbf{Precision} & \textbf{Recall} & \textbf{Pearson Corr} 
& \textbf{Accuracy} & \textbf{F1-Score} & \textbf{Precision} & \textbf{Recall} \\

\hline
\multicolumn{10}{l}{\textbf{Encoder-Only}} \\
\hline
BERT\cite{devlin-etal-2019-bert} &39.11&38.24&42.06&39.92&16.90 & \underline{30.89} &\underline{27.78}&\underline{31.78}&27.95 \\
DistilBERT\cite{sanh2019distilbert} &36.89 &36.16 &39.79 &37.73 & 14.23 &30.11&23.77&28.09&25.06 \\
mBERT & \textbf{42.22} & \textbf{42.37} & \underline{44.59} & \textbf{42.58} & \textbf{25.14} &29.44&27.08&28.84&27.88 \\
XLM-RoBERTa\cite{conneau2019unsupervised} &\underline{40.67}&\underline{40.81} &42.50&\underline{40.90} & \underline{22.28} &30.27&23.65&27.09&26.06 \\
DeBERTa\cite{he2020deberta} &39.00&37.73&44.01&40.25 & 19.04 &30.11&20.95&25.16&24.60 \\
ALBERT\cite{lan2019albert} &36.89 &34.91 &41.99 &38.25& 16.10&30.59 &25.78&\textbf{32.78}&\textbf{28.95} \\
XLNet\cite{yang2019xlnet} &36.78&35.99&39.48&37.59 &13.00 &29.56&25.79&30.38&26.25 \\
\hline
\multicolumn{10}{l}{\textbf{Encoder-Decoder}} \\
\hline
BART\cite{lewis-etal-2020-bart} &38.11&36.62&41.78&39.30& 13.98 &29.44&25.35&28.53&25.80 \\
mBART\cite{liu2020mbart} &30.11&15.43&10.04&33.33& 13.00 &\textbf{31.56}&\textbf{28.94}&29.89&\underline
{28.60} \\
T5\cite{JMLR:v21:20-074} &37.89&36.88&41.58&39.07 &14.88 &28.78&20.11&26.51&23.33 \\
\hline
\multicolumn{10}{l}{\textbf{Decoder-Only (Zero-Shot)}} \\
\hline
Gemma-2-2b-it\cite{team2024gemma} &33.00&27.66&37.61&35.20& 13.47&21.56&10.16&11.15&17.76 \\
Llama-3.2-3b-it\cite{touvron2023llama} &38.44&33.07&\textbf{46.08}&39.57 &15.23 &24.89&18.44&22.96&20.43\\
\hline
\multicolumn{10}{l}{\textbf{Decoder-Only (Few-Shot)}} \\
\hline
Gemma-2-2b-it &36.00&27.80&23.15&35.04& 15.55 &22.56&14.87&15.75&19.97 \\
Llama-3.2-3b-it &34.78&30.27&41.22&35.28 &13.79 &19.89&16.55&18.88&20.65\\
\hline
\multicolumn{10}{l}{\textbf{Decoder-Only (QLoRA Fine-Tuned)}} \\
\hline
Gemma-2-2b-it &37.22&20.49&28.02&26.09& 13.72 &16.89&9.31&17.41&19.00 \\
Llama-3.2-3b-it &20.67&9.60&14.22&12.38&15.25 &23.11&14.42&16.62&17.54 \\
\hline
\end{tabular}
}
\caption{Baseline Metrics of Models Across Satire Levels and Humor Attributes. The top results are represented in \textbf{bold}, and the second-best results are \underline{underlined}.}

\label{tab:baseline_results}
\end{table*}

\section{Human Evaluation} \label{humaneval}
To ensure data quality and reliability, we manually annotated the test set (900 samples) of DHD for human evaluation. Labeling deceptive humor is challenging and requires a deep understanding of fabricated claims and context. We trained five annotators using detailed guidelines and conducted a mock annotation round to identify and clarify ambiguous cases related to overlapping humor types and fabricated claims. After providing personalized feedback and ensuring annotators had a clear understanding of the task and claims, we proceeded with the final annotation phase. \textcolor{blue}{\autoref{tab:agreement}} shows the agreement between human and machine labels. For the Satire Level, unweighted and weighted Cohen’s Kappa scores\cite{cohen1960coefficient, cohen1968weighted} were calculated. Weighted Kappa accounts for partial agreement by penalizing distant mismatches. Fair unweighted (20\% - 40\%) and moderately weighted (40\% - 60\%) agreement is observed in most languages. For Humor Attributes, moderate agreement indicates good alignment with human perception, with English showing substantial agreement, reflecting better clarity.

\begin{table}[H]
\resizebox{0.48\textwidth}{!}{
\begin{tabular}{lccc}
\hline
\textbf{Lang} & \textbf{Satire Unwt K} & \textbf{Satire Wtd K} & \textbf{Humor Unwt K} \\
\hline
English & 44.97 & 55.30 & 62.52 \\
Telugu* & 37.67 & 42.90 & 45.11 \\
Hindi* & 34.50 & 43.56 & 48.63 \\
Kannada* & 33.37 & 37.99 & 40.94 \\
Tamil* & 33.78 & 40.27 & 46.00 \\
\hline
\textbf{Overall} & 35.98 & 43.07 & 47.38 \\
\hline
\end{tabular}
}
\caption{Human-Machine Alignment. Agreement metrics: Unwt K = unweighted Cohen's Kappa; Wtd K = weighted Cohen's Kappa.}
\label{tab:agreement}
\end{table}

We observed fair to moderate agreement for DHD, which is expected, given the complex nature of deceptive humor. Satire Level labels are highly subjective, making exact agreement challenging. Additionally, some Humor Attribute labels, such as Absurdity, Wordplay, and sometimes Irony, can overlap or be used interchangeably, further complicating consistent labeling. While LLMs generally generate high-quality comments, internal biases can occasionally produce unclear or hard-to-interpret outputs. In our test set, about 15 out of 900 such comments were identified and removed. Conversely, even when LLM-generated comments are accurate, human annotators might misinterpret them if they miss the hidden intent or layered meaning behind the humor. These challenges reflect the inherent difficulty in understanding and labeling deceptive humor.

To assess the quality of the synthetic Deceptive Humor data across different languages, we evaluated three core aspects: \textit{Readability}, \textit{Claim-Graspability}, and \textit{Cultural Nuance}. Readability captures both the grammatical correctness and linguistic fluency of the generated content. Claim-Graspability measures whether humans can intuitively understand the hidden claim or narrative embedded in the comment, a crucial property for deceptive content. Finally, Cultural Nuance evaluates whether the humor felt organically human or if it exhibited artificial patterns, indicating machine generation. As shown in \textcolor{blue}{\autoref{tab:quality_asses}}, English comments ranked highest across all criteria, while languages like Kannada and Tamil showed slightly lower scores, especially in Cultural Nuance, likely due to limited pretraining exposure and culturally rooted humor gaps. 

\begin{table}[H]
\resizebox{0.48\textwidth}{!}{
\begin{tabular}{lccc}
\hline
\textbf{Lang} & \textbf{Readability} & \textbf{Claim-Graspability} & \textbf{Cultural Nuance} \\
\hline
English & 9.00 & 7.80 & 8.20 \\
Telugu*  & 7.90 & 7.60 & 7.10 \\
Hindi*   & 7.70 & 7.20 & 7.20 \\
Kannada* & 7.10 & 6.30 & 6.20 \\
Tamil*   & 7.30 & 6.50 & 5.90 \\
\hline
\end{tabular}
}
\caption{Quality Assessment for Deceptive Humor Data (scale of 1–10; 1 = low, 10 = high).}
\label{tab:quality_asses}
\end{table}

\section{Error Analysis}\label{error_analysis}
This section highlights examples where the models misclassified Satire Level and Humor Attribute. We examine the original labels, predictions, and possible reasons for these errors. These insights help uncover recurring patterns of confusion, especially in detecting indirect satire and subtle humor constructs.\\

\noindent \textbf{Satire Level:}\\
Here we present misclassified samples where comments originally labeled as High Satire (level 3) were predicted as Low Satire (level 1). This highlights the challenge models face in detecting subtle or indirect forms of strong satire.

\begin{tcolorbox}[title=Example 1: Satire Level]
\textbf{Comment:} Are temples now just stepping stones for interfaith harmony campaigns? \\
\textbf{Original Label:} 3 (High Satire) \\
\textbf{Predicted:} 1 (Low Satire) \\
\textbf{Possible model interpretation:} The model likely misjudged because the satire is indirect and culturally nuanced. \\
\textbf{Conclusion:} Indirect satire, especially on sensitive topics (religion etc), is challenging for the model to detect accurately.
\end{tcolorbox}

\begin{tcolorbox}[title=Example 2: Satire Level]
\textbf{Comment:} Seems like nud*ty is the new immunity booster. \\
\textbf{Original Label:} 3 (High Satire) \\
\textbf{Predicted:} 1 (Low Satire) \\
\textbf{Possible model interpretation:} The indirect form of satire caused confusion in prediction. \\
\textbf{Conclusion:} The model underestimates satire when it is subtle or implied rather than explicit.
\end{tcolorbox}

\noindent \textbf{Humor Attribute:}\\ 
This section shows examples of humor attribute misclassifications where the model confused one humor type for another. Such errors underscore the difficulty in distinguishing nuanced humor styles like irony, wordplay, and absurdity.

\begin{tcolorbox}[title=Example 1: Humor Attribute]
\textbf{Comment:} Ch*na's virus: the only war fought with sweatpants and Wi-Fi! \\
\textbf{Original Label:} Absurdity \\
\textbf{Predicted:} Irony \\
\textbf{Possible model interpretation:} The contrast between “war” and “sweatpants/Wi-Fi” appeared ironic due to the unexpected pairing. \\
\textbf{Conclusion:} Although clearly absurd, LLMs confuse extreme exaggeration with irony, revealing challenges in humor nuance detection.
\end{tcolorbox}

\begin{tcolorbox}[title=Example 2: Humor Attribute]
\textbf{Comment:} So the gas cylinder decided to skip being cooked on and went straight for a track record? \\
\textbf{Original Label:} Wordplay \\
\textbf{Predicted:} Irony \\
\textbf{Possible model interpretation:} The phrasing suggested ironic intent rather than recognizing the pun. \\
\textbf{Conclusion:} Clever wordplay misread as irony, highlighting difficulty in identifying linguistic constructs.
\end{tcolorbox}

\section{Ethical Consideration}
Due to the sensitive and potentially misleading nature of content in the DHD, the dataset must be used strictly for research purposes. The dataset includes humorous content that may embed misinformation, satire, or deceptive cues, which could be misinterpreted or misused outside of a controlled academic setting. Access will be granted only to researchers who formally agree to use the dataset responsibly and ethically. The primary goal of DHD is to support the development and evaluation of computational models capable of detecting and understanding deceptive humor. Any commercial use, public redistribution, or application with potential societal harm is strictly prohibited.
\section{Conclusion and Future work}\label{conclusion}
In this study, we proposed a novel research direction at the intersection of misinformation and humorous content, emphasizing how humor can act as a major vehicle for spreading misinformation. This approach highlighted the need to better understand the interplay between humor and misinformation, which is often overlooked in existing research. The study underscored the importance of addressing this issue as it plays a significant role in shaping public perception and influencing societal narratives. To support this exploration, we introduce the DHD dataset along with strong baselines to guide future research. We hope this work encourages deeper inquiry into humor-driven misinformation and inspires innovative solutions in this emerging domain. Ultimately, our findings aim to catalyze responsible AI development in understanding and mitigating the subtle threats posed by deceptive humor.


\bibliographystyle{ACM-Reference-Format}
\bibliography{sample-base}


\begin{thebibliography}{36}


\ifx \showCODEN    \undefined \def \showCODEN     #1{\unskip}     \fi
\ifx \showISBNx    \undefined \def \showISBNx     #1{\unskip}     \fi
\ifx \showISBNxiii \undefined \def \showISBNxiii  #1{\unskip}     \fi
\ifx \showISSN     \undefined \def \showISSN      #1{\unskip}     \fi
\ifx \showLCCN     \undefined \def \showLCCN      #1{\unskip}     \fi
\ifx \shownote     \undefined \def \shownote      #1{#1}          \fi
\ifx \showarticletitle \undefined \def \showarticletitle #1{#1}   \fi
\ifx \showURL      \undefined \def \showURL       {\relax}        \fi
\providecommand\bibfield[2]{#2}
\providecommand\bibinfo[2]{#2}
\providecommand\natexlab[1]{#1}
\providecommand\showeprint[2][]{arXiv:#2}

\bibitem[Bhardwaj et~al\mbox{.}(2020)]%
        {bhardwaj2020hostility}
\bibfield{author}{\bibinfo{person}{Mohit Bhardwaj}, \bibinfo{person}{Md~Shad
  Akhtar}, \bibinfo{person}{Asif Ekbal}, \bibinfo{person}{Amitava Das}, {and}
  \bibinfo{person}{Tanmoy Chakraborty}.} \bibinfo{year}{2020}\natexlab{}.
\newblock \showarticletitle{Hostility detection dataset in Hindi}.
\newblock \bibinfo{journal}{\emph{arXiv preprint arXiv:2011.03588}}
  (\bibinfo{year}{2020}).
\newblock


\bibitem[Biradar et~al\mbox{.}(2024)]%
        {biradar2024faux}
\bibfield{author}{\bibinfo{person}{Shankar Biradar}, \bibinfo{person}{Sunil
  Saumya}, {and} \bibinfo{person}{Arun Chauhan}.}
  \bibinfo{year}{2024}\natexlab{}.
\newblock \showarticletitle{Faux Hate: unravelling the web of fake narratives
  in spreading hateful stories: a multi-label and multi-class dataset in
  cross-lingual Hindi-English code-mixed text}.
\newblock \bibinfo{journal}{\emph{Language Resources and Evaluation}}
  (\bibinfo{year}{2024}), \bibinfo{pages}{1--32}.
\newblock


\bibitem[Brown et~al\mbox{.}(2020)]%
        {brown2020language}
\bibfield{author}{\bibinfo{person}{Tom Brown}, \bibinfo{person}{Benjamin Mann},
  \bibinfo{person}{Nick Ryder}, \bibinfo{person}{Melanie Subbiah},
  \bibinfo{person}{Jared~D Kaplan}, \bibinfo{person}{Prafulla Dhariwal},
  \bibinfo{person}{Arvind Neelakantan}, \bibinfo{person}{Pranav Shyam},
  \bibinfo{person}{Girish Sastry}, \bibinfo{person}{Amanda Askell},
  {et~al\mbox{.}}} \bibinfo{year}{2020}\natexlab{}.
\newblock \showarticletitle{Language models are few-shot learners}.
\newblock \bibinfo{journal}{\emph{Advances in neural information processing
  systems}}  \bibinfo{volume}{33} (\bibinfo{year}{2020}),
  \bibinfo{pages}{1877--1901}.
\newblock


\bibitem[Cohen(1960)]%
        {cohen1960coefficient}
\bibfield{author}{\bibinfo{person}{Jacob Cohen}.}
  \bibinfo{year}{1960}\natexlab{}.
\newblock \showarticletitle{A coefficient of agreement for nominal scales}.
\newblock \bibinfo{journal}{\emph{Educational and psychological measurement}}
  \bibinfo{volume}{20}, \bibinfo{number}{1} (\bibinfo{year}{1960}),
  \bibinfo{pages}{37--46}.
\newblock


\bibitem[Cohen(1968)]%
        {cohen1968weighted}
\bibfield{author}{\bibinfo{person}{Jacob Cohen}.}
  \bibinfo{year}{1968}\natexlab{}.
\newblock \showarticletitle{Weighted kappa: Nominal scale agreement provision
  for scaled disagreement or partial credit.}
\newblock \bibinfo{journal}{\emph{Psychological bulletin}}
  \bibinfo{volume}{70}, \bibinfo{number}{4} (\bibinfo{year}{1968}),
  \bibinfo{pages}{213}.
\newblock


\bibitem[Conneau et~al\mbox{.}(2019)]%
        {conneau2019unsupervised}
\bibfield{author}{\bibinfo{person}{Alexis Conneau}, \bibinfo{person}{Kartikay
  Khandelwal}, \bibinfo{person}{Naman Goyal}, \bibinfo{person}{Vishrav
  Chaudhary}, \bibinfo{person}{Guillaume Wenzek}, \bibinfo{person}{Francisco
  Guzm{\'a}n}, \bibinfo{person}{Edouard Grave}, \bibinfo{person}{Myle Ott},
  \bibinfo{person}{Luke Zettlemoyer}, {and} \bibinfo{person}{Veselin
  Stoyanov}.} \bibinfo{year}{2019}\natexlab{}.
\newblock \showarticletitle{Unsupervised cross-lingual representation learning
  at scale}.
\newblock \bibinfo{journal}{\emph{arXiv preprint arXiv:1911.02116}}
  (\bibinfo{year}{2019}).
\newblock


\bibitem[Dettmers et~al\mbox{.}(2024)]%
        {dettmers2024qlora}
\bibfield{author}{\bibinfo{person}{Tim Dettmers}, \bibinfo{person}{Artidoro
  Pagnoni}, \bibinfo{person}{Ari Holtzman}, {and} \bibinfo{person}{Luke
  Zettlemoyer}.} \bibinfo{year}{2024}\natexlab{}.
\newblock \showarticletitle{Qlora: Efficient finetuning of quantized llms}.
\newblock \bibinfo{journal}{\emph{Advances in Neural Information Processing
  Systems}}  \bibinfo{volume}{36} (\bibinfo{year}{2024}).
\newblock


\bibitem[Devlin et~al\mbox{.}(2019)]%
        {devlin-etal-2019-bert}
\bibfield{author}{\bibinfo{person}{Jacob Devlin}, \bibinfo{person}{Ming-Wei
  Chang}, \bibinfo{person}{Kenton Lee}, {and} \bibinfo{person}{Kristina
  Toutanova}.} \bibinfo{year}{2019}\natexlab{}.
\newblock \showarticletitle{{BERT}: Pre-training of Deep Bidirectional
  Transformers for Language Understanding}. In
  \bibinfo{booktitle}{\emph{Proceedings of the 2019 Conference of the North
  {A}merican Chapter of the Association for Computational Linguistics: Human
  Language Technologies, Volume 1 (Long and Short Papers)}},
  \bibfield{editor}{\bibinfo{person}{Jill Burstein}, \bibinfo{person}{Christy
  Doran}, {and} \bibinfo{person}{Thamar Solorio}} (Eds.).
  \bibinfo{publisher}{Association for Computational Linguistics},
  \bibinfo{address}{Minneapolis, Minnesota}, \bibinfo{pages}{4171--4186}.
\newblock
\href{https://doi.org/10.18653/v1/N19-1423}{doi:\nolinkurl{10.18653/v1/N19-1423}}


\bibitem[Gaspar et~al\mbox{.}(2023)]%
        {gaspar2023laughter}
\bibfield{author}{\bibinfo{person}{Methasani-Redona Gaspar, Joseph~P}
  {et~al\mbox{.}}} \bibinfo{year}{2023}\natexlab{}.
\newblock \showarticletitle{Laughter and Lies: Unraveling the Intricacies of
  Humor and Deception}.
\newblock \bibinfo{journal}{\emph{Current Opinion in Psychology}}
  (\bibinfo{year}{2023}), \bibinfo{pages}{101707}.
\newblock


\bibitem[Gautam et~al\mbox{.}(2024)]%
        {gautam2024blind}
\bibfield{author}{\bibinfo{person}{Srinath~Mukund Gautam, Sanjana}
  {et~al\mbox{.}}} \bibinfo{year}{2024}\natexlab{}.
\newblock \showarticletitle{Blind Spots and Biases: Exploring the Role of
  Annotator Cognitive Biases in NLP}. In \bibinfo{booktitle}{\emph{Proceedings
  of the Third Workshop on Bridging Human--Computer Interaction and Natural
  Language Processing}}. \bibinfo{publisher}{Association for Computational
  Linguistics}, \bibinfo{address}{Mexico City, Mexico},
  \bibinfo{pages}{82--88}.
\newblock
\href{https://doi.org/10.18653/v1/2024.hcinlp-1.8}{doi:\nolinkurl{10.18653/v1/2024.hcinlp-1.8}}


\bibitem[He et~al\mbox{.}(2020)]%
        {he2020deberta}
\bibfield{author}{\bibinfo{person}{Pengcheng He}, \bibinfo{person}{Xiaodong
  Liu}, \bibinfo{person}{Jianfeng Gao}, {and} \bibinfo{person}{Weizhu Chen}.}
  \bibinfo{year}{2020}\natexlab{}.
\newblock \showarticletitle{Deberta: Decoding-enhanced bert with disentangled
  attention}.
\newblock \bibinfo{journal}{\emph{arXiv preprint arXiv:2006.03654}}
  (\bibinfo{year}{2020}).
\newblock


\bibitem[Horvitz et~al\mbox{.}(2024)]%
        {horvitz2024getting}
\bibfield{author}{\bibinfo{person}{Zachary Horvitz}, \bibinfo{person}{Jingru
  Chen}, \bibinfo{person}{Rahul Aditya}, \bibinfo{person}{Harshvardhan
  Srivastava}, \bibinfo{person}{Robert West}, \bibinfo{person}{Zhou Yu}, {and}
  \bibinfo{person}{Kathleen McKeown}.} \bibinfo{year}{2024}\natexlab{}.
\newblock \showarticletitle{Getting Serious about Humor: Crafting Humor
  Datasets with Unfunny Large Language Models}.
\newblock \bibinfo{journal}{\emph{arXiv preprint arXiv:2403.00794}}
  (\bibinfo{year}{2024}).
\newblock


\bibitem[Hossain et~al\mbox{.}(2019)]%
        {hossain2019president}
\bibfield{author}{\bibinfo{person}{Nabil Hossain}, \bibinfo{person}{John
  Krumm}, {and} \bibinfo{person}{Michael Gamon}.}
  \bibinfo{year}{2019}\natexlab{}.
\newblock \showarticletitle{" President Vows to Cut< Taxes> Hair": Dataset and
  Analysis of Creative Text Editing for Humorous Headlines}.
\newblock \bibinfo{journal}{\emph{arXiv preprint arXiv:1906.00274}}
  (\bibinfo{year}{2019}).
\newblock


\bibitem[Hurst et~al\mbox{.}(2024)]%
        {hurst2024gpt}
\bibfield{author}{\bibinfo{person}{Aaron Hurst}, \bibinfo{person}{Adam Lerer},
  \bibinfo{person}{Adam~P Goucher}, \bibinfo{person}{Adam Perelman},
  \bibinfo{person}{Aditya Ramesh}, \bibinfo{person}{Aidan Clark},
  \bibinfo{person}{AJ Ostrow}, \bibinfo{person}{Akila Welihinda},
  \bibinfo{person}{Alan Hayes}, \bibinfo{person}{Alec Radford},
  {et~al\mbox{.}}} \bibinfo{year}{2024}\natexlab{}.
\newblock \showarticletitle{Gpt-4o system card}.
\newblock \bibinfo{journal}{\emph{arXiv preprint arXiv:2410.21276}}
  (\bibinfo{year}{2024}).
\newblock


\bibitem[Joshi et~al\mbox{.}(2015)]%
        {joshi2015harnessing}
\bibfield{author}{\bibinfo{person}{Aditya Joshi}, \bibinfo{person}{Vinita
  Sharma}, {and} \bibinfo{person}{Pushpak Bhattacharyya}.}
  \bibinfo{year}{2015}\natexlab{}.
\newblock \showarticletitle{Harnessing context incongruity for sarcasm
  detection}. In \bibinfo{booktitle}{\emph{Proceedings of the 53rd Annual
  Meeting of the Association for Computational Linguistics and the 7th
  International Joint Conference on Natural Language Processing (Volume 2:
  Short Papers)}}. \bibinfo{pages}{757--762}.
\newblock


\bibitem[Kenton and Toutanova(2019)]%
        {kenton2019bert}
\bibfield{author}{\bibinfo{person}{Jacob Devlin Ming-Wei~Chang Kenton} {and}
  \bibinfo{person}{Lee~Kristina Toutanova}.} \bibinfo{year}{2019}\natexlab{}.
\newblock \showarticletitle{Bert: Pre-training of deep bidirectional
  transformers for language understanding}. In
  \bibinfo{booktitle}{\emph{Proceedings of naacL-HLT}},
  Vol.~\bibinfo{volume}{1}. Minneapolis, Minnesota.
\newblock


\bibitem[Lan et~al\mbox{.}(2019)]%
        {lan2019albert}
\bibfield{author}{\bibinfo{person}{Zhenzhong Lan}, \bibinfo{person}{Mingda
  Chen}, \bibinfo{person}{Sebastian Goodman}, \bibinfo{person}{Kevin Gimpel},
  \bibinfo{person}{Piyush Sharma}, {and} \bibinfo{person}{Radu Soricut}.}
  \bibinfo{year}{2019}\natexlab{}.
\newblock \showarticletitle{Albert: A lite bert for self-supervised learning of
  language representations}.
\newblock \bibinfo{journal}{\emph{arXiv preprint arXiv:1909.11942}}
  (\bibinfo{year}{2019}).
\newblock


\bibitem[Lewis et~al\mbox{.}(2020)]%
        {lewis-etal-2020-bart}
\bibfield{author}{\bibinfo{person}{Mike Lewis}, \bibinfo{person}{Yinhan Liu},
  \bibinfo{person}{Naman Goyal}, \bibinfo{person}{Marjan Ghazvininejad},
  \bibinfo{person}{Abdelrahman Mohamed}, \bibinfo{person}{Omer Levy},
  \bibinfo{person}{Veselin Stoyanov}, {and} \bibinfo{person}{Luke
  Zettlemoyer}.} \bibinfo{year}{2020}\natexlab{}.
\newblock \showarticletitle{{BART}: Denoising Sequence-to-Sequence Pre-training
  for Natural Language Generation, Translation, and Comprehension}. In
  \bibinfo{booktitle}{\emph{Proceedings of the 58th Annual Meeting of the
  Association for Computational Linguistics}},
  \bibfield{editor}{\bibinfo{person}{Dan Jurafsky}, \bibinfo{person}{Joyce
  Chai}, \bibinfo{person}{Natalie Schluter}, {and} \bibinfo{person}{Joel
  Tetreault}} (Eds.). \bibinfo{publisher}{Association for Computational
  Linguistics}, \bibinfo{address}{Online}, \bibinfo{pages}{7871--7880}.
\newblock
\href{https://doi.org/10.18653/v1/2020.acl-main.703}{doi:\nolinkurl{10.18653/v1/2020.acl-main.703}}


\bibitem[Liu et~al\mbox{.}(2024)]%
        {liu2024best}
\bibfield{author}{\bibinfo{person}{Ruibo Liu}, \bibinfo{person}{Jerry Wei},
  \bibinfo{person}{Fangyu Liu}, \bibinfo{person}{Chenglei Si},
  \bibinfo{person}{Yanzhe Zhang}, \bibinfo{person}{Jinmeng Rao},
  \bibinfo{person}{Steven Zheng}, \bibinfo{person}{Daiyi Peng},
  \bibinfo{person}{Diyi Yang}, \bibinfo{person}{Denny Zhou}, {et~al\mbox{.}}}
  \bibinfo{year}{2024}\natexlab{}.
\newblock \showarticletitle{Best practices and lessons learned on synthetic
  data for language models}.
\newblock \bibinfo{journal}{\emph{arXiv preprint arXiv:2404.07503}}
  (\bibinfo{year}{2024}).
\newblock


\bibitem[Liu et~al\mbox{.}(2020)]%
        {liu2020mbart}
\bibfield{author}{\bibinfo{person}{Yinhan Liu}, \bibinfo{person}{Jiatao Gu},
  \bibinfo{person}{Naman Goyal}, \bibinfo{person}{Xian Li},
  \bibinfo{person}{Sergey Edunov}, \bibinfo{person}{Marjan Ghazvininejad},
  \bibinfo{person}{Mike Lewis}, {and} \bibinfo{person}{Luke Zettlemoyer}.}
  \bibinfo{year}{2020}\natexlab{}.
\newblock \showarticletitle{Multilingual Denoising Pre-training for Neural
  Machine Translation}.
\newblock \bibinfo{journal}{\emph{arXiv preprint arXiv:2001.08210}}
  (\bibinfo{year}{2020}).
\newblock


\bibitem[Potash et~al\mbox{.}(2017)]%
        {potash2017semeval}
\bibfield{author}{\bibinfo{person}{Peter Potash}, \bibinfo{person}{Alexey
  Romanov}, {and} \bibinfo{person}{Anna Rumshisky}.}
  \bibinfo{year}{2017}\natexlab{}.
\newblock \showarticletitle{Semeval-2017 task 6:\# hashtagwars: Learning a
  sense of humor}. In \bibinfo{booktitle}{\emph{Proceedings of the 11th
  International Workshop on Semantic Evaluation (SemEval-2017)}}.
  \bibinfo{pages}{49--57}.
\newblock


\bibitem[Raffel et~al\mbox{.}(2020)]%
        {JMLR:v21:20-074}
\bibfield{author}{\bibinfo{person}{Colin Raffel}, \bibinfo{person}{Noam
  Shazeer}, \bibinfo{person}{Adam Roberts}, \bibinfo{person}{Katherine Lee},
  \bibinfo{person}{Sharan Narang}, \bibinfo{person}{Michael Matena},
  \bibinfo{person}{Yanqi Zhou}, \bibinfo{person}{Wei Li}, {and}
  \bibinfo{person}{Peter~J. Liu}.} \bibinfo{year}{2020}\natexlab{}.
\newblock \showarticletitle{Exploring the Limits of Transfer Learning with a
  Unified Text-to-Text Transformer}.
\newblock \bibinfo{journal}{\emph{Journal of Machine Learning Research}}
  \bibinfo{volume}{21}, \bibinfo{number}{140} (\bibinfo{year}{2020}),
  \bibinfo{pages}{1--67}.
\newblock
\urldef\tempurl%
\url{http://jmlr.org/papers/v21/20-074.html}
\showURL{%
\tempurl}


\bibitem[Rubin et~al\mbox{.}(2016)]%
        {rubin2016fake}
\bibfield{author}{\bibinfo{person}{Victoria~L Rubin}, \bibinfo{person}{Niall
  Conroy}, \bibinfo{person}{Yimin Chen}, {and} \bibinfo{person}{Sarah
  Cornwell}.} \bibinfo{year}{2016}\natexlab{}.
\newblock \showarticletitle{Fake news or truth? using satirical cues to detect
  potentially misleading news}. In \bibinfo{booktitle}{\emph{Proceedings of the
  second workshop on computational approaches to deception detection}}.
  \bibinfo{pages}{7--17}.
\newblock


\bibitem[Sanh et~al\mbox{.}(2019)]%
        {sanh2019distilbert}
\bibfield{author}{\bibinfo{person}{Victor Sanh}, \bibinfo{person}{Lysandre
  Debut}, \bibinfo{person}{Julien Chaumond}, {and} \bibinfo{person}{Thomas
  Wolf}.} \bibinfo{year}{2019}\natexlab{}.
\newblock \showarticletitle{DistilBERT, a distilled version of BERT: smaller,
  faster, cheaper and lighter}.
\newblock \bibinfo{journal}{\emph{arXiv preprint arXiv:1910.01108}}
  (\bibinfo{year}{2019}).
\newblock


\bibitem[Shabbir et~al\mbox{.}(2007)]%
        {shabbir2007use}
\bibfield{author}{\bibinfo{person}{Thwaites~Des Shabbir, Haseeb}
  {et~al\mbox{.}}} \bibinfo{year}{2007}\natexlab{}.
\newblock \showarticletitle{The use of humor to mask deceptive advertising:
  It's no laughing matter}.
\newblock \bibinfo{journal}{\emph{Journal of Advertising}}
  \bibinfo{volume}{36}, \bibinfo{number}{2} (\bibinfo{year}{2007}),
  \bibinfo{pages}{75--85}.
\newblock


\bibitem[Sharma et~al\mbox{.}(2020)]%
        {sharma-etal-2020-semeval}
\bibfield{author}{\bibinfo{person}{Chhavi Sharma}, \bibinfo{person}{Deepesh
  Bhageria}, \bibinfo{person}{William Scott}, \bibinfo{person}{Srinivas PYKL},
  \bibinfo{person}{Amitava Das}, \bibinfo{person}{Tanmoy Chakraborty},
  \bibinfo{person}{Viswanath Pulabaigari}, {and} \bibinfo{person}{Bj{\"o}rn
  Gamb{\"a}ck}.} \bibinfo{year}{2020}\natexlab{}.
\newblock \showarticletitle{{S}em{E}val-2020 Task 8: Memotion Analysis- the
  Visuo-Lingual Metaphor!}. In \bibinfo{booktitle}{\emph{Proceedings of the
  Fourteenth Workshop on Semantic Evaluation}},
  \bibfield{editor}{\bibinfo{person}{Aurelie Herbelot},
  \bibinfo{person}{Xiaodan Zhu}, \bibinfo{person}{Alexis Palmer},
  \bibinfo{person}{Nathan Schneider}, \bibinfo{person}{Jonathan May}, {and}
  \bibinfo{person}{Ekaterina Shutova}} (Eds.).
  \bibinfo{publisher}{International Committee for Computational Linguistics},
  \bibinfo{address}{Barcelona (online)}, \bibinfo{pages}{759--773}.
\newblock
\href{https://doi.org/10.18653/v1/2020.semeval-1.99}{doi:\nolinkurl{10.18653/v1/2020.semeval-1.99}}


\bibitem[Team et~al\mbox{.}(2023)]%
        {team2023gemini}
\bibfield{author}{\bibinfo{person}{Gemini Team}, \bibinfo{person}{Rohan Anil},
  \bibinfo{person}{Sebastian Borgeaud}, \bibinfo{person}{Jean-Baptiste
  Alayrac}, \bibinfo{person}{Jiahui Yu}, \bibinfo{person}{Radu Soricut},
  \bibinfo{person}{Johan Schalkwyk}, \bibinfo{person}{Andrew~M Dai},
  \bibinfo{person}{Anja Hauth}, \bibinfo{person}{Katie Millican},
  {et~al\mbox{.}}} \bibinfo{year}{2023}\natexlab{}.
\newblock \showarticletitle{Gemini: a family of highly capable multimodal
  models}.
\newblock \bibinfo{journal}{\emph{arXiv preprint arXiv:2312.11805}}
  (\bibinfo{year}{2023}).
\newblock


\bibitem[Team et~al\mbox{.}(2024)]%
        {team2024gemma}
\bibfield{author}{\bibinfo{person}{Gemma Team}, \bibinfo{person}{Thomas
  Mesnard}, \bibinfo{person}{Cassidy Hardin}, \bibinfo{person}{Robert Dadashi},
  \bibinfo{person}{Surya Bhupatiraju}, \bibinfo{person}{Shreya Pathak},
  \bibinfo{person}{Laurent Sifre}, \bibinfo{person}{Morgane Rivi{\`e}re},
  \bibinfo{person}{Mihir~Sanjay Kale}, \bibinfo{person}{Juliette Love},
  {et~al\mbox{.}}} \bibinfo{year}{2024}\natexlab{}.
\newblock \showarticletitle{Gemma: Open models based on gemini research and
  technology}.
\newblock \bibinfo{journal}{\emph{arXiv preprint arXiv:2403.08295}}
  (\bibinfo{year}{2024}).
\newblock


\bibitem[Thorne et~al\mbox{.}(2018)]%
        {thorne2018fever}
\bibfield{author}{\bibinfo{person}{James Thorne}, \bibinfo{person}{Andreas
  Vlachos}, \bibinfo{person}{Christos Christodoulopoulos}, {and}
  \bibinfo{person}{Arpit Mittal}.} \bibinfo{year}{2018}\natexlab{}.
\newblock \showarticletitle{FEVER: a large-scale dataset for fact extraction
  and VERification}.
\newblock \bibinfo{journal}{\emph{arXiv preprint arXiv:1803.05355}}
  (\bibinfo{year}{2018}).
\newblock


\bibitem[Touvron et~al\mbox{.}(2023)]%
        {touvron2023llama}
\bibfield{author}{\bibinfo{person}{Hugo Touvron}, \bibinfo{person}{Thibaut
  Lavril}, \bibinfo{person}{Gautier Izacard}, \bibinfo{person}{Xavier
  Martinet}, \bibinfo{person}{Marie-Anne Lachaux},
  \bibinfo{person}{Timoth{\'e}e Lacroix}, \bibinfo{person}{Baptiste
  Rozi{\`e}re}, \bibinfo{person}{Naman Goyal}, \bibinfo{person}{Eric Hambro},
  \bibinfo{person}{Faisal Azhar}, {et~al\mbox{.}}}
  \bibinfo{year}{2023}\natexlab{}.
\newblock \showarticletitle{Llama: Open and efficient foundation language
  models}.
\newblock \bibinfo{journal}{\emph{arXiv preprint arXiv:2302.13971}}
  (\bibinfo{year}{2023}).
\newblock


\bibitem[Van~Hee et~al\mbox{.}(2018)]%
        {van2018semeval}
\bibfield{author}{\bibinfo{person}{Cynthia Van~Hee}, \bibinfo{person}{Els
  Lefever}, {and} \bibinfo{person}{V{\'e}ronique Hoste}.}
  \bibinfo{year}{2018}\natexlab{}.
\newblock \showarticletitle{Semeval-2018 task 3: Irony detection in english
  tweets}. In \bibinfo{booktitle}{\emph{Proceedings of the 12th international
  workshop on semantic evaluation}}. \bibinfo{pages}{39--50}.
\newblock


\bibitem[Wang et~al\mbox{.}(2017)]%
        {wang2017liar}
\bibfield{author}{\bibinfo{person}{William~Yang Wang} {et~al\mbox{.}}}
  \bibinfo{year}{2017}\natexlab{}.
\newblock \showarticletitle{" liar, liar pants on fire": A new benchmark
  dataset for fake news detection}.
\newblock \bibinfo{journal}{\emph{arXiv preprint arXiv:1705.00648}}
  (\bibinfo{year}{2017}).
\newblock


\bibitem[Wu et~al\mbox{.}(2023)]%
        {wu2023synthetic}
\bibfield{author}{\bibinfo{person}{Yawen Wu}, \bibinfo{person}{Zhepeng Wang},
  \bibinfo{person}{Dewen Zeng}, \bibinfo{person}{Yiyu Shi}, {and}
  \bibinfo{person}{Jingtong Hu}.} \bibinfo{year}{2023}\natexlab{}.
\newblock \showarticletitle{Synthetic data can also teach: Synthesizing
  effective data for unsupervised visual representation learning}. In
  \bibinfo{booktitle}{\emph{Proceedings of the AAAI Conference on Artificial
  Intelligence}}, Vol.~\bibinfo{volume}{37}. \bibinfo{pages}{2866--2874}.
\newblock


\bibitem[Xu et~al\mbox{.}(2023)]%
        {xu2023s2ynre}
\bibfield{author}{\bibinfo{person}{Benfeng Xu}, \bibinfo{person}{Quan Wang},
  \bibinfo{person}{Yajuan Lyu}, \bibinfo{person}{Dai Dai},
  \bibinfo{person}{Yongdong Zhang}, {and} \bibinfo{person}{Zhendong Mao}.}
  \bibinfo{year}{2023}\natexlab{}.
\newblock \showarticletitle{S2ynRE: Two-stage self-training with synthetic data
  for low-resource relation extraction}. In
  \bibinfo{booktitle}{\emph{Proceedings of the 61st Annual Meeting of the
  Association for Computational Linguistics (Volume 1: Long Papers)}}.
  \bibinfo{pages}{8186--8207}.
\newblock


\bibitem[Yang et~al\mbox{.}(2019)]%
        {yang2019xlnet}
\bibfield{author}{\bibinfo{person}{Zhilin Yang}, \bibinfo{person}{Zihang Dai},
  \bibinfo{person}{Yiming Yang}, \bibinfo{person}{Jaime Carbonell},
  \bibinfo{person}{Russ~R Salakhutdinov}, {and} \bibinfo{person}{Quoc~V Le}.}
  \bibinfo{year}{2019}\natexlab{}.
\newblock \showarticletitle{Xlnet: Generalized autoregressive pretraining for
  language understanding}.
\newblock \bibinfo{journal}{\emph{Advances in neural information processing
  systems}}  \bibinfo{volume}{32} (\bibinfo{year}{2019}).
\newblock


\bibitem[Zhou et~al\mbox{.}(2020)]%
        {zhou2020survey}
\bibfield{author}{\bibinfo{person}{Zafarani~Reza Zhou, Xinyi} {et~al\mbox{.}}}
  \bibinfo{year}{2020}\natexlab{}.
\newblock \showarticletitle{A survey of fake news: Fundamental theories,
  detection methods, and opportunities}.
\newblock \bibinfo{journal}{\emph{ACM Computing Surveys (CSUR)}}
  \bibinfo{volume}{53}, \bibinfo{number}{5} (\bibinfo{year}{2020}),
  \bibinfo{pages}{1--40}.
\newblock


\end{thebibliography}
 


\end{document}